\documentclass{article}

\usepackage{spconf}

\usepackage{color, colortbl}
\definecolor{Gray}{gray}{0.85}
\usepackage{url}
\usepackage{multirow}
\usepackage{hhline}
\usepackage[pdftex]{graphicx}

\usepackage{array}
\newcolumntype{C}{>{\centering\arraybackslash} m{1.35cm} }

\title{Descriptors and regions of interest fusion for gender classification 
in the wild. Comparison and combination with Convolutional Neural Networks}

\name{M. Castrill\'on-Santana, J. Lorenzo-Navarro and E. Ram\'on-Balmaseda\thanks{Work partially funded by the Institute of Intelligent Systems and Numerical Applications in Engineering and the Computer Science Department at ULPGC.}}
\address{SIANI - Universidad de Las Palmas de Gran Canaria\\Spain}%

\begin{document}

\maketitle

\begin{abstract}
Gender classification (GC) has achieved high accuracy in different experimental evaluations based mostly on
inner facial details. However, these results do not generalize well in unrestricted datasets and particularly in cross-database experiments, where the
performance drops drastically. In this paper, we analyze the state-of-the-art GC accuracy on three large datasets: MORPH, LFW and GROUPS. 
We discuss their respective difficulties and bias, concluding that the most challenging and wildest complexity is present in GROUPS. This 
dataset covers hard conditions such as low resolution imagery and cluttered background. 
Firstly, we analyze in depth the performance of different descriptors extracted 
from the face and its local context on this dataset. Selecting the bests 
and studying their most suitable combination allows us to design a
solution that beats any previously published results for GROUPS with the Dago's 
protocol, reaching an accuracy 
over $94.2\%$, reducing the gap with other simpler datasets.
The chosen solution based on local descriptors is later evaluated in a 
cross-database scenario with the three 
mentioned datasets, and full dataset 5-fold cross validation. 
The achieved results are compared with a Convolutional Neural Network approach,
achieving rather similar marks.
Finally, a solution is proposed combining both focuses, exhibiting great 
complementarity, boosting GC performance to beat 
previously published results
in GC both cross-database, and full in-database evaluations. 
\end{abstract}

\begin{keywords}
Gender classification, HOG, LBP, LSP, LOSIB, information fusion, face local context, cross-database, CNN
\end{keywords}

\section{Introduction}


Gender is a feature easily extracted by humans, and quite useful for human interaction. After all, gender classification (GC) is not yet a solved problem
for the computer vision community. Automatic GC is nowadays an active research field in computer 
vision, with different application scenarios covering surveillance, demographics or direct marketing among others.

\begin{figure}[!t]  \begin{minipage}[t]{0.9\linewidth}
    \centering
   
    \begin{minipage}[c]{0.73\linewidth}
      \centering
      \includegraphics[width=0.9\linewidth]{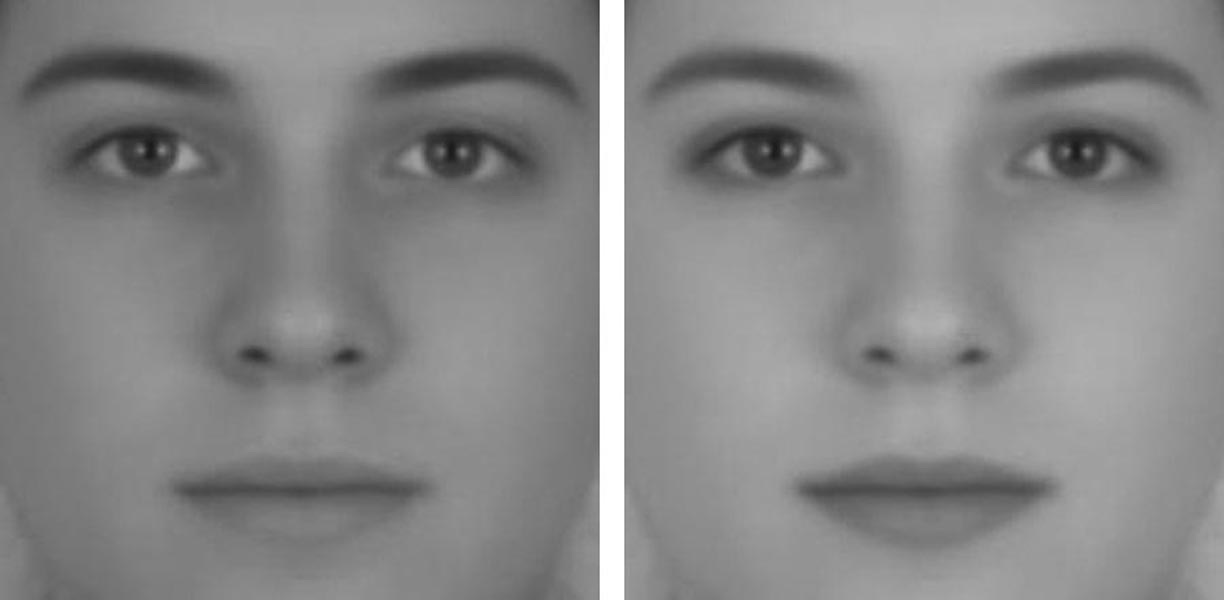}\\
      (a)
     \end{minipage}\hfill 
  \begin{minipage}[c]{0.26\linewidth}
      \centering
      \includegraphics[width=\linewidth]{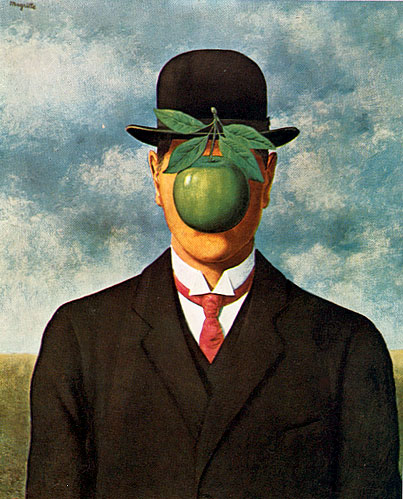}\\
      (b)
    \end{minipage}\hfill 
   
  \end{minipage}
\caption{(a) Images extracted from~\cite{Russell09-p}. Both images are obtained from the same original, but their contrast has been altered, 
for most human observers the face on the left appears male, while the face on the right
appears female. (b) Le Fils de l'homme (Ren\'e Magritte). \label{fig:Paintings}}

\end{figure}

Nowadays, state-of-the-art GC approaches achieve relative high performance based just on visual facial features. However, the exclusive 
observation of the face might produce perception errors and restrict scenarios of application. 
In the former, the face pattern may create perception illusions~\cite{Russell09-p}, see Figure~\ref{fig:Paintings}a. 
The latter occurs in low resolution applications, such as surveillance scenarios, where the facial local context may help to estimate the gender
of an individual, see Figure~\ref{fig:Paintings}b.

 \begin{table}[ht]
\caption{Cross-database accuracy rates in the literature evaluating large iage collections: $^1$ inter-ocular distance $> 20$, $^2$ $>20$ years old , 
$^3$ automatically detected faces of $>20$ years old, $^4$ single face per identity,
$^5$ LFW subset containing 10147 samples.}\label{tab:cross-database}
\centering
\begin{footnotesize}
\centering
 \begin{tabular}{c|c|c|c}
  Reference & Training set& Test set& Accuracy (\%) \\ \hline
  \cite{Bekios11-pami}& FERET & UCN & $81.29$ \\
  \cite{Bekios11-pami}& PAL & UCN & $74.09$ \\
    \cite{Ramon12-ciarp}& MORPH & LFW &  $75.10$ \\
  \cite{Erdogmus14-mmsp} & MORPH &  LFW &  $76.64$ \\ 
  \cite{Dago11-befit}& GROUPS$^1$ & LFW &  $89.77$  \\
    \cite{Mansanet16-prl} & GROUPS$^1$ & LFW & $94.48$\\
  \cite{Bekios14-prl}& GROUPS$^2$ & LFW & $79.53$ \\
  \cite{Danisman14-icpr}& GROUPS$^3$ & LFW &  $91.62$ \\
  \cite{Danisman15-esa}& GROUPS$^3$ & LFW$^4$ &  $93.35$ \\
  \cite{JiaS15-prl}& 4 million faces & LFW &  $96.86$ \\
  \cite{Antipov15-prl} & CASIA WebFace~\cite{YiD14-arxiv} & LFW$^5$& $97.10$ \\
  \cite{Ramon12-ciarp}& MORPH & GROUPS &  $76.74$ \\
  \cite{Dago11-befit}& LFW & GROUPS$^1$ &  $81.02$ \\
  \cite{Mansanet16-prl}& LFW & GROUPS$^1$ &  $83.03$ \\
  \cite{Danisman15-esa}& LFW & GROUPS$^3$ &  $85.00$ \\
  
\end{tabular}
\end{footnotesize}
\label{cross-databaselit}
\end{table}

\subsection{Related work}

In this paper, a first objective is to analyze GC results of different datasets to identify those that are closer to real scenarios.
As mentioned above, most state-of-the-art GC approaches focus on the facial pattern. This 
is evidenced by the latest problem surveys~\cite{NgCB15-paa,Ngan15-nist}, and recent results in major 
journals~\cite{Bekios11-pami,Danisman15-esa,JiaS15-prl,Alexandre10-prl,Makinen08-pami,Tapia13-ifs,Andreu14-ivc}.

As in any classification problem, GC accuracy is estimated on different datasets obtaining the so-called in-database accuracy. 
However, researchers
must not just be interested in getting higher and stable recognition
rates for a particularly database, but also improving cross-database classification 
rates, i.e.
testing with an independent dataset, whose images were captured with different 
conditions. Cross-database classification is closer to real situations 
with no dataset bias, that has been proven to provoke 
optimistic accuracies~\cite{Bekios14-prl,Baluja07-ijcv}. 
Indeed, in real scenarios a gender 
classifier is trained with a set of images, and later deployed under conditions 
that may differ from those of the training dataset. 
Table~\ref{tab:cross-database} summarizes recent cross-database classification rates 
of large databases. 
We claim that high performance can be obtained particularly in 
homogeneous, biased and/or reduced datasets, with good image quality, and restricted poses. 
To illustrate this, different experiments 
on FERET~\cite{Phillips98-ivc} have recently reported very high GC rates~\cite{Bekios11-pami,Alexandre10-prl,Makinen08-pami}.
However, classifiers trained with FERET
do not perform robustly with other datasets. The performance drops notoriously~\cite{Andreu14-ivc}. Observing Table~\ref{tab:cross-database}, with 
the exception of testing with the biased LFW dataset,
the accuracy hardly reaches $80\%$, evidencing a lack of accuracy 
in cross-database classification. . 

Therefore, part of the community is currently addressing a more realistic or general problem, i.e. GC 
in the wild. Thus,
researchers are now giving more attention to experiments with newer and larger databases that enclose more variability 
in terms of 1) identity, age and ethnicity, 2) pose and illumination conditions, and 3) image resolution.




Focusing on large and heterogeneous datasets, we highlight recent results reported
for the non public UCN~\cite{Bekios11-pami}, and the available  MORPH \cite{Ricanek06-fgc}, 
GROUPS~\cite{Gallagher09-cvpr}, and LFW~\cite{huang07-tr} datasets. 
Observing the in-database rates in Table~\ref{in-databaselit}, there is not much room for improvement for LFW and MORPH. 
We argue that both datasets present some level of simplification that benefits the overall accuracy achieved. In fact, both include multiple samples of the same identity, circumstance that clearly mix gender and
identity classification. 
On the other side, GROUPS offers a less restricted scenario, reporting the lowest 
accuracy, with a large gap compared to other datasets. This evidence has convinced 
us to  focus on this particular dataset, agreeing with the 2015 NIST
report conclusion on the topic~\cite{Ngan15-nist}. We aim at reducing this GC accuracy gap.




\begin{table}[!t]
\caption{Recent in-database accuracy results in large datasets. $^1$ inter ocular distance $>20$, $^3$ $22778$ aut. detected faces, $^4$~$7443$ of the total $13233$ images, $^5$ BEFIT protocol, 
$^6$ balanced subset with $14244$ of the total $55134$ images.}
\centering
\begin{footnotesize}
\centering
 \begin{tabular}{c|c|c}
  Reference & Dataset& Accuracy (\%) \\ \hline
  \cite{Shan12-prl} &  LFW$^4$ &  $94.81$ \\
  \cite{Tapia13-ifs} &  LFW$^4$ & $98.01$ \\ 
  \cite{Dago11-befit} &  LFW$^5$ &  $97.23$ \\ 
  \cite{Mansanet16-prl} &LFW$^5$ & $96.25$ \\ 
  \cite{ElDin13-lncs} & LFW & $91.5$ \\
   \cite{Dago11-befit} &  GROUPS$^1$ &  $84.55-86.61$ \\ 
   \cite{Castrillon13-ciarp} & GROUPS$^1$ & $88.1$ \\
   \cite{Mansanet16-prl} &GROUPS$^1$ & $91.59$ \\
   \cite{Castrillon16-prl} & GROUPS$^1$ & $92.46$ \\
     \cite{Kumar11-pami} & GROUPS$^3$ &  $86.4$  \\
     \cite{Bekios14-prl} & GROUPS$^3$ &  $80.5$  \\     
  \cite{ChenH14-pami} & GROUPS$^3$ &  $90.4$ \\
  \cite{Chu10-icpr} & MORPH &  $88$\\
  \cite{Ramon12-ciarp} & MORPH$^6$ & $97.1$\\
\end{tabular}
\end{footnotesize}
\label{in-databaselit}
\end{table}

Certainly, the need of facial features restricts the context of application, requiring a visible and almost frontal face. 
From another point of view, different researchers have recently investigated the inclusion of external 
facial features~\cite{Kumar11-pami,Bourdev11-iccv} such as
~hair, clothing~\cite{ChenH12-eccv,Li12-n,Freire14-icip}, 
their combination with other cues~\cite{Hadid09-pr}, or even features extracted from 
the body~\cite{Collins09-iccvw,Guo09-accv}. 
The latter claims to be better adapted to real surveillance scenarios
where the facial pattern is noisy, not frontal, occluded, or presents low resolution. However, 
their application is particularly restricted, as no body occlusion may be present. 

Indeed, the inclusion of non facial features is consistent with the human vision system that employs 
external and other features for GC, such as gait, body contours, hair, clothing, 
voice, etc.~\cite{Kumar11-pami,Toseeb12-pone}. 
These considerations seem to be of particular interest for 
degraded, low resolution or noisy images~\cite{Sinha96-nature}. 
For those reasons, we will include in our study features extracted from the face, and its local context.

\subsection{Contributions}

Summarizing, the paper contributions in relation to recent literature are: 
1) a study of 
state-of-the-art GC accuracies in large datasets, analyzing
the elements that characterize the current problem challenges, 
2) an extensive
experimental analysis of a wide collection of descriptors in GROUPS, identifying the best local 
descriptors for the problem, and an analysis of
their robustness against noise; 
3) the reduction of the accuracy gap compared to other datasets making use of a score fusion architecture combining
features and regions of interest, outperforming previous results, while confirming 
that both descriptors and regions provide complementary information for the problem; 
7) a further analysis of GC failures for GROUPS;
4) the translation of the approach to cross-database scenarios;
 including GROUPS, LFW and MORPH. 
5) the comparison with a CNN architecture,
6) and the proposal of a solution that combines local descriptors and CNN, boosting
significantly previous results;

\section{Representation and classification}
\label{sec:faces}

\begin{figure}[tb]
\centering
\includegraphics[width=0.65\linewidth]{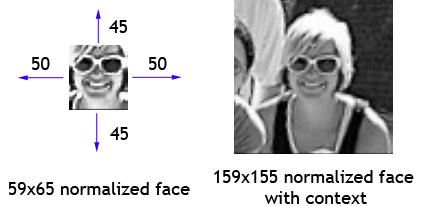}
\caption{Normalized face (F) and its corresponding face with context (HS). Each face image is rotated, 
re-scaled and cropped so that the center of each eye is placed at a fixed
location (pixels (16,17) and (42,17) for F). Sample from GROUPS.}
\label{fig:sample-facesandctx}
\end{figure}

Figure~\ref{fig:sample-facesandctx} illustrates on the left a typical face pattern used for GC. To get 
this image, a normalization step rotates, scales and crops the image to fix the eyes in specific locations and inter-eye distance ($26$ pixels), with the resolution of $59 \times 65$ pixels. This pattern is referred below as the face (F).

On the right in 
Figure~\ref{fig:sample-facesandctx}, both the face and its local context are presented. The pattern encloses the hair, shoulders, 
part of the upper chest, and some background. This is the pattern that we refer hereafter 
as head and shoulders (HS).
The final $159 \times 155$ pattern exhibits a large resolution for a real surveillance scenario, and increases the number of features.
For that reason and to bring the resolutions closer to those found in real scenarios, 
in the local descriptors experiments we have considered the HS pattern 
scaled down to $64\times 64$, $32 \times 32$ and $16 \times 16$ pixels.
Accordingly, in those lower resolution images, the aproximated inter-eye distances are respectively $10$, $5$ and $3$ pixels. 

\subsection{Representation}

Local pixel-wise descriptors have recently received lots of attention with several successful applications to facial analysis~\cite{Ahonen06-pami,Deniz11-prl}.
~We briefly describe those included in the experimental study.

\subsubsection{Histograms of Oriented Gradients (HOG)}

\begin{figure}[hbt]  
\begin{minipage}[t]{0.9\linewidth}
    \centering
    \centering
    \begin{minipage}[c]{0.6\linewidth}
      \centering
      \includegraphics[scale=0.5]{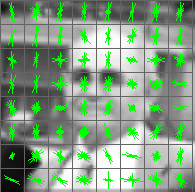}\\
    \end{minipage}\hfill
     \begin{minipage}[c]{0.4\linewidth}
      \centering
      \includegraphics[scale=0.5]{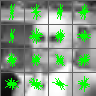}\\
    \end{minipage}\hfill 
    
  \end{minipage}
 \caption{Relative size of the different HS pattern resolutions considered: respectively $64 \times 64$ and $32 \times 32$. Their respective HOG grid is depicted.
     \label{fig:Tomas}}
     \end{figure}

HOG~\cite{Dalal05-cvpr} is based on the histogram of 
the gradient orientations in a regular area of the image, called cell. The image is divided into cells, concatenating their respective histograms to compose the descriptor. In order to reduce the illumination influence, each histogram is normalized 
using a neighborhood, called block. In their original implementation~\cite{Dalal05-cvpr}, the cell size is $8 \times 8$ pixels and the block is $2 \times 2$ cells. 
The configuration parameters are the number of bins in the histogram, the angle range $0-180^{\circ}$ or $0-360^{\circ}$, the norm used in the normalization 
stage inside the block, and the overlapping between blocks in the image. 
In the experiments, we used a $8 \times 8$ cells grid, and $9$ bins following the implementation by Ludwig et al.~\cite{Ludwig09-itsc}. 
Figure~\ref{fig:Tomas} illustrates the gradient orientation 
 in each cell of the image, for a sample at different resolutions. 

\subsubsection{Local Binary Patterns (LBP)}

LBP~\cite{Ojala02-pami} is a robust and simple but efficient texture descriptor that labels the pixels of an image by thresholding the 
pixel neighborhood. 
In texture classification, the LBP code occurrences in an image are described using a histogram. 

However, for facial recognition this approximation implies the loss of spatial 
information.
~The alternative proposed in~\cite{Ahonen06-pami}, 
divides the face into 
small regions where the LBP operator is applied and later concatenated
~into a 
single histogram. 
The textures of the facial regions are locally encoded, 
and the combination of these micro-patterns histograms generates a comprehensive description of the face image.

An extension reduces the dictionary of LBP codes observing the most common ones in texture images. Uniform LBP, $LBP^{u2}$, contains at most two bitwise transitions 
from $0$ to $1$ or vice versa.
NILBP~\cite{LiuL12-ivc} is another LBP variant that tries to reduce the LBP local structure oversimplification, computing the difference with 
the neighbors mean, $\mu$, instead of the central pixel gray value.

A recent redefinition known as Local Salient Patterns (LSP)~\cite{Chai13-icb} focuses on the location of the largest differences within 
the pixel neighborhood. LSP has reported better rates in identity recognition. 

Finally, LOSIB~\cite{Olalla14-icpr} is a descriptor enhancer based on LBP that computes local oriented statistical information in the whole image. 
We have adapted it to face analysis concatenating the histograms obtained from a grid of cells.

The chosen operators are studied below applied on the normalized facial pattern (F) divided into a grid of $n \times n$ cells, in our case, $n=5$. 



\subsection{Classification}

As the focus of this paper is in the combination of descriptors more than the classifiers themselves, we have used the well known Support Vector Machine (SVM) 
classifier with RBF kernel~\cite{Vapnik95}. 
As the reader knows,
the SVM classifier obtains the hyperplane that maximizes the class separation to minimize the 
risk. For non linear separable problems, a previous mapping of the original feature space to a higher dimensional one is carried out 
by means of kernels. 


\section{Databases}
\label{sec:databases}

\begin{figure}[t]  
\begin{minipage}[t]{1.0\linewidth}
    \centering
   \centering
    \begin{minipage}[c]{0.43\linewidth}
      \centering
      \includegraphics[scale=0.46]{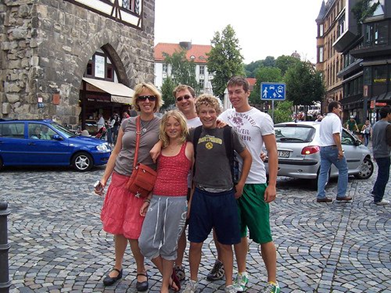}
    \end{minipage}\hfill 
     \begin{minipage}[c]{0.29\linewidth}
      \centering
      \includegraphics[scale=0.4]{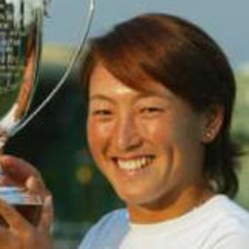}
    \end{minipage}\hfill
    \begin{minipage}[c]{0.28\linewidth}
      \centering
      \includegraphics[scale=0.35]{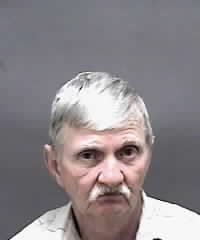}
    \end{minipage}\hfill        \\ 
     \caption{Sample images respectively of GROUPS, LFW and MORPH.
     Their respective original resolutions are $391 \times 293$, $249 \times 249$  
     and $200 \times 240$ pixels, suggesting a relevant difference in the 
     facial pattern resolution.
     \label{fig:DBSamples}}
  \end{minipage}
\end{figure}

We argued above that an experimental setup with a small or restricted database is not representative for a real world scenario where the gender classifier 
must face up with thousands of people. To overcome this limitation, we have selected three public
databases with a large number of individuals acquired, with the aim at including 
larger variability. A sample of each dataset in shown in Figure~\ref{fig:DBSamples}, 
and their respective statistics are presented in 
Table~\ref{tabla-datasets}, summarizing their main features as follows:
\begin{itemize}
\item \textbf{MORPH}~\cite{Ricanek06-fgc}. This set contains images of more 
than $13,000$ identities. 
We observe however, three undesired features for in the wild scenarios: 1) the number of samples per 
class is not balanced, 2) the images were 
acquired indoor in 
rather similar resolution and illumination conditions (capture 
bias~\cite{Torralba11-cvpr}), and 3) there are multiple samples per individual.

\item \textbf{Labeled Faces in the Wild} (\textbf{LFW})~\cite{huang07-tr}. The dataset includes images of 
$5,749$ individuals captured under less controlled conditions.  
~However, 1) it contains
several samples per individual, 2) the number of samples per class is not balanced, and 3) the inclusion of public people introduce a selection bias~\cite{Torralba11-cvpr}.

\item \textbf{The Images of Groups} (\textbf{GROUPS}). This database~\cite{Gallagher09-cvpr} 
contains more than $28,000$ labeled faces of lower resolution. According to Table~\ref{in-databaselit}, and the recent 
FRVT report~\cite{Ngan15-nist}, this 
database is the hardest for GC. Observing Table~\ref{tabla-datasets}, this is explained due to the lower face resolution ($IE$) and the larger out of 
plane rotations ($\bar{\sigma}_{EN}$).

\end{itemize}

\begin{table}
\caption{Databases characteristics: Instances (per class), SNR, inter eye distance 
mean ($IE$), and normalized standard deviation of the eye-nose distance ($\bar{\sigma}_{EN}$).}
\begin{footnotesize}
\begin{center}
\begin{tabular}{l|c|c|c|c}
Database & Total (female/male) & SNR & $IE$ & $\bar{\sigma}_{EN}$\\
\hline GROUPS & 28,220 (14,549/13,671) & 36.91 &25.32  $\pm$ 15.4& 0.50  \\
\hline LFW & 13,232 (2,970/10,252)& 36.08& 42.16  $\pm$ 4.5& 0.15\\
\hline MORPH & 55,134 (8,488/46,646)& 31.44& 92.97  $\pm$ 26.4& 0.2\\
\end{tabular} 
\end{center}
\end{footnotesize}
\label{tabla-datasets}
\end{table}


\section{Experiments}


As mentioned before, we firstly focus on GROUPS to analyze the accuracy achieved by a wide collection of descriptors. 
For this aim, we have followed the experimental protocol defined by Dago et al.~\cite{Dago11-befit}. This protocol defines a 5-fold cross-validation 
experimental setup that contains the subset of faces automatically
detected with an inter-eye distance larger than $20$ 
pixels\footnote{Available at BEFIT site, visit 
http://i14s50.anthropomatik.kit.edu/431.php}. 
Secondly, we define 
strategies to improve accuracy and analyze incorrectly classified
samples to improve the overall accuracy. Later, aiming at 
verifying the proposal generalization, we design an in- and cross-database 
experimental setup
that considers the full three selected datasets: GROUPS, LFW and MORPH. Those 
results are also
compared with a CNN designed for GC, to finally conclude the great benefits that
the combination of local descriptors and CNN offer to boost GC performance.

\subsection{GROUPS Dago's protocol}\label{sec:Dago} 

The results achieved for the Dago's protocol are summarized in Table~\ref{tab:Dagoresultsbasics}. The experiments have covered
the local descriptors collection described above.

The table includes results using both the facial (F), and the head and shoulders (HS) 
patterns. We remind the reader that F
has a resolution of $59 \times 65$ pixels, and three different resolutions 
have been used for HS: 
$16\times16$, $32\times32$ and $64\times64$, see Section~\ref{sec:faces} 
for more details. 
Observe that the facial resolution contained in HS is up to eight times lower than in F.

For F we employed as descriptors HOG, LBP$^{u2}$, NILBP, LOSIB, and LSP histograms (respectively F-HOG, F-LBP$^{u2}$, F-NILBP, F-LOSIB and F-LSP). For HOG, we have 
selected $8 \times 8$ cell histograms with $9$ bins. 
For the rest of descriptors, we have made 
use of $5 \times 5 $ histograms, attending our previous experience in~\cite{Ramon12-ciarp}. HS is described in terms of 
HOG and LOSIB features, but considering for the former different resolutions of the pattern. 

As baseline, we have included classifiers trained with the first $100$ PCA components obtained from the original normalized gray facial images, the histograms obtained from the facial pattern using LBP and HOG, and HS using HOG (F-PCA, F-HOG-PCA, F-LBP-PCA, and HS$_{64 \times 64}$-HOG-PCA).
For classification we present results for SVM+RBF with C (trade-off between margin and error) and gamma values, respectively tuned in the range of $C=[0.25,8]$ and $gamma=[0.04,0.15]$. 

The best two accuracies for F are provided by: F-HOG (C1), F-LBP (C3), while the 
best one for HS is reported by HS$_{64 \times 64}$-HOG (C2). 
In particular the representation based on F-HOG
beats most literature rates of this dataset~\cite{Dago11-befit}, including 
our previous results with linear kernels~\cite{Castrillon13-ciarp}. That 
descriptor reached $88.23\%$. 
It is also remarkable the accuracy achieved using the lowest HS resolution, $75.31\%$.  Observe that 
Dago et al.~\cite{Dago11-befit} made use of a pattern almost twice larger, and the number of features in our previous work 
was significantly larger. These accuracies were found significantly 
different ($p=4\cdot 10 ^{-4}$) after carrying out a Kruskal-Wallis test because a previous Jarque-Bera test ($p=0.5$) rejected the normality of the samples.

The table includes the classification rates per gender, a detail commonly skipped in 
the literature. In most cases the female accuracy is slightly lower,
similarly to the conclusions of the 2015 NIST report~\cite{Ngan15-nist}.

\begin{table}[ht]
\caption{GC accuracy (in brackets per class: female/male) achieved using different sets of features ($n$ number of features), resolutions and patterns (F and HS)
following the Dago's protocol. Processing time in milliseconds.}
\centering
\begin{footnotesize}
 \begin{tabular}{c|c|c|c}
   \multirow{2}{*}{Pattern-Descriptor} & \multirow{2}{*}{Accuracy (\%)} & \multirow{2}{*}{n}  & Proc. \\ 
     &  &  &  time\\ 
  \hline  
  F-PCA& $77.91$  ($77.86$/$77.95$) & 100 & $1.2$ \\
  \hline
  F-HOG (\textbf{C1}) & $\textbf{88.23}$ ($88.20$/$88.25$) &  576 & $5.7$ \\ 
  \hline
  F-HOG-PCA & $81.10$  ($81.02$/$81.38$) &100 & $1$ \\  
  \hline
  F-LBP$^{u2}$ (\textbf{C3})& $86.74$ ($86.29$/$87.20$)  & 1475 & $48.3$ \\ 
   \hline 
  F-LBP$^{u2}$-PCA & $80.45$  ($80.15$/$80.76$)& 100 & $1$ \\ 
  \hline
  F-NILBP & $85.31$  ($85.02$/$85.59$) & 1425 & $48.3$ \\  
  \hline
  F-LOSIB (\textbf{C4}) & $86.65$ ($86.00$/$87.31$)  & 512 & $10.4$ \\ 
  \hline
  F-LSP$^{0}$ & $85.58$ ($84.98$/$81.17$)  & 1425 & $39.7$ \\ 
  \hline  
  F-LSP$^{1}$ & $85.27$ ($84.85$/$85.69$)  & 1425 & $31.3$ \\ 
  \hline  
  F-LSP$^{2}$ & $82.92$ ($81.94$/$83.91$) & 1425 & $31.3$ \\  
  \hline
  \hline
  HS$_{64\times 64}$-HOG-PCA & $80.80$  ($80.15$/$80.76$)  & 100 & $1$ \\
  \hline 
  HS$_{64\times 64}$-HOG (\textbf{C2})& $\textbf{85.93}$  ($83.69$/$88.11$) & 576& $6.2$ \\ 
     \hline
  HS$_{64\times 64}$-LOSIB (\textbf{C5}) & $82.72$ ($81.41$/$84.06$)  & 512 & $10.4$ \\  
  \hline
  HS$_{32\times 32}$-HOG &$85.04$  ($83.13$/$86.99$) & 576&  $11.3$\\ 
   \hline  
  HS$_{16\times 16}$-HOG& $75.31$ ($75.08$/$75.56$) & 576& $18.5$ \\  
  
\end{tabular}
\end{footnotesize}
\label{tab:Dagoresultsbasics}
\end{table}

\begin{figure}[t]  \begin{minipage}[t]{1.0\linewidth}
  \centering
      \includegraphics[width=0.75\linewidth]{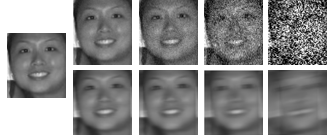}
 \caption{Original normalized face ($59 \times 65$), and resulting images after 
 applying Gaussian (first row ) or 
 Blurring (second row) noise with different 
 magnitudes: a variance for the gaussian noise up to $0.1$, and a 
 linear motion up to 21 pixels for blurring.}\label{fig:noiseaccuracyimgs}
  \end{minipage}
  
\end{figure}

\subsection{Robustness against noise}

An important element to study, in low resolution scenarios, is the robustness of the proposed approach to the presence of noise. GROUPS contains faces of different resolutions. Of the total number of samples, a $5\%$ of them 
present an inter-eye distance lower than $10$ pixels, and $41\%$ lower than $20$ pixels.
Focusing on lower resolution, 
we include
an additional experiment introducing noise to avoid the advantage of getting low resolution images down-sampling higher resolution patterns.
Thus, we have noised the images before extracting HOG features. 

\begin{figure}[t]  
\begin{minipage}[t]{1.0\linewidth}
    \centering
   

      \begin{minipage}[t]{1.0\linewidth}
    \centering   
    \begin{minipage}[c]{0.5\linewidth}
      \centering
      \includegraphics[width=\linewidth]{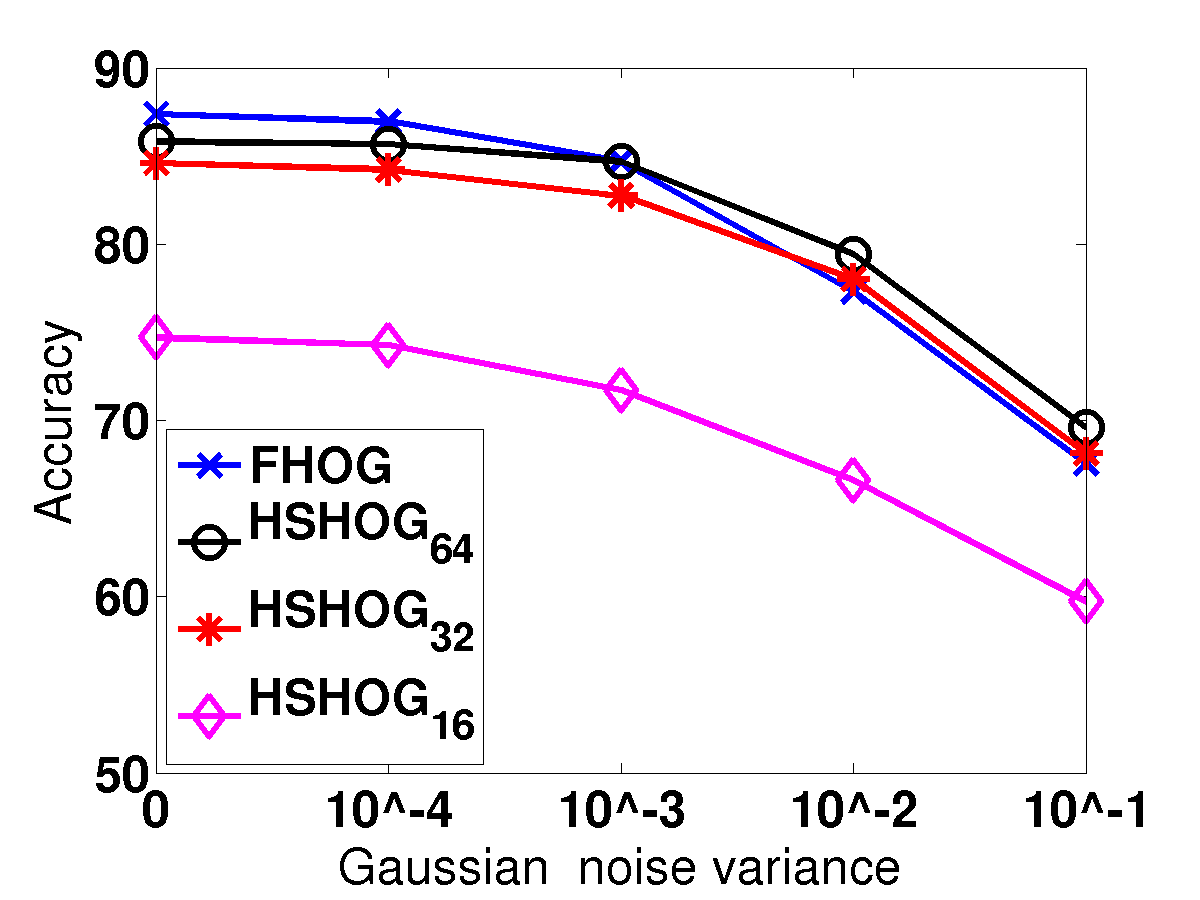}
      (a)
     \end{minipage}\hfill 
  \begin{minipage}[c]{0.5\linewidth}
      \centering
      \includegraphics[width=\linewidth]{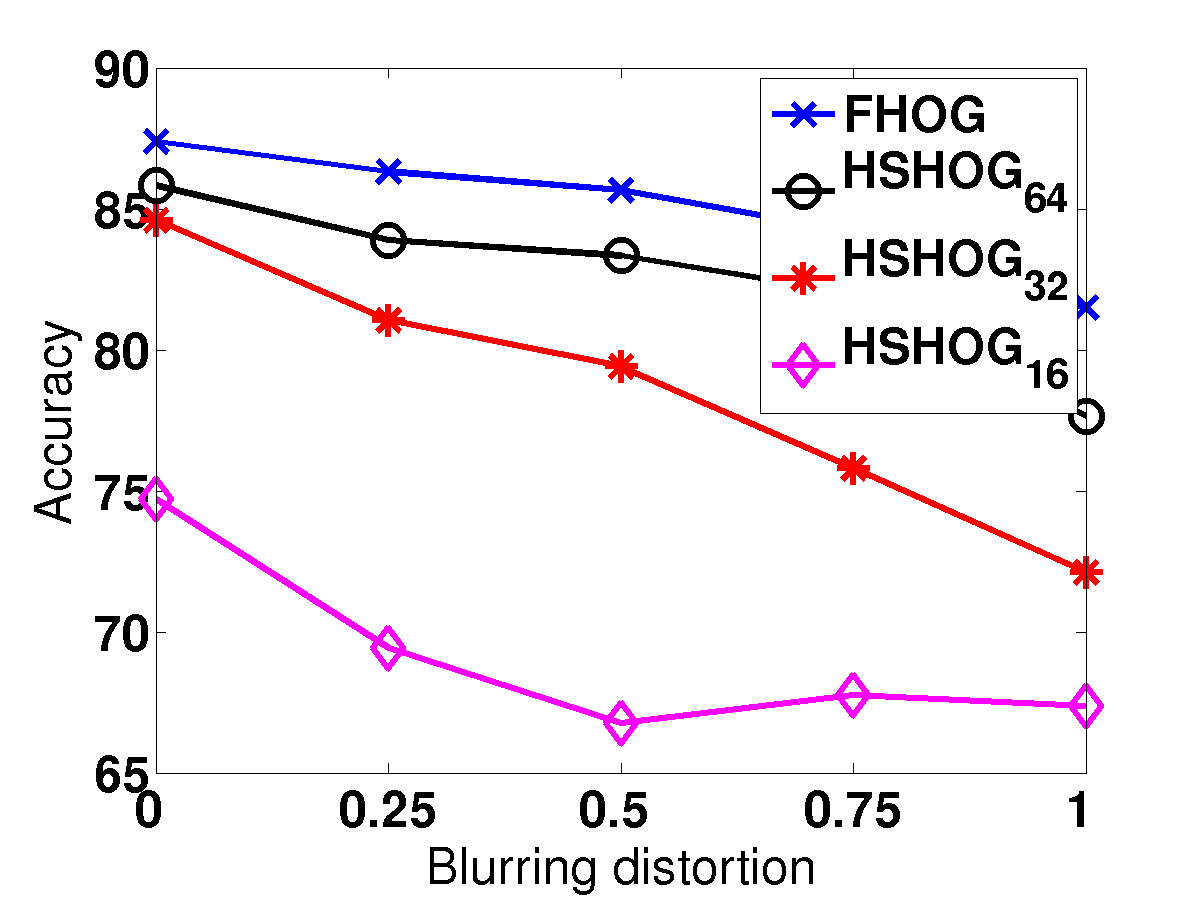}
      (b)
    \end{minipage}\hfill 
  \end{minipage}

\caption{Accuracies with (a) Gaussian and (b) blur noise.\label{fig:noiseaccuracy} }
  \end{minipage}
  
\end{figure}

In Figure~\ref{fig:noiseaccuracyimgs} we present the original and resulting patterns 
after applying noise of different nature and magnitude. Figures~\ref{fig:noiseaccuracy}a-b reports the noise influence in GC 
comparing different pattern resolutions.
The accuracy achieved considering only the face pattern (F) is largely affected by the Gaussian noise; reporting lower accuracy than for HS when the 
noise magnitude increases, see Figure~\ref{fig:noiseaccuracy}a.

\subsection{Ensemble of classifiers}

Due to the nature of the different descriptors studied, they might provide complementary information. 
Thus, the combination of all of them in a stacking fashion~\cite{Wolpert92} can improve the overall performance, added to the evidences 
in the reduction of ambiguous cases occurrences~\cite{Castrillon15-iciap}. 

\begin{figure}[tb]  
    \centering
\begin{minipage}[t]{1\linewidth}
    \centering
   \includegraphics[width=\linewidth]{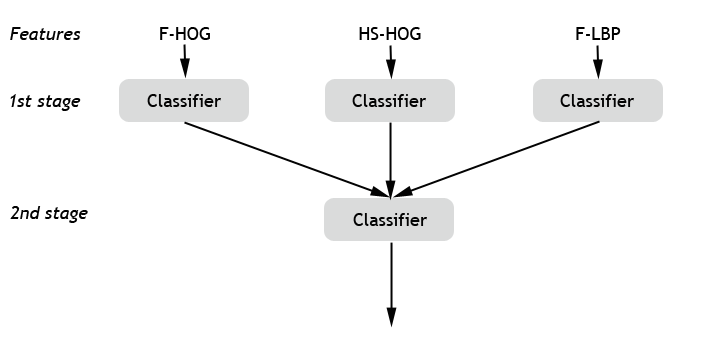}
   \caption{Illustration of an stacked classifiers architecture, with three classifiers in the first stage whose scores are fed into a 
   second stage "`meta"' classifier. \label{fig:Stacking}}
  \end{minipage}
\end{figure}

We explore below score level fusion, as feature level fusion did not report a notorious accuracy improvement, requiring to evaluate a much more complex multi-feature problem.
Therefore, we have followed a two stage stacking architecture as illustrated in Figure~\ref{fig:Stacking}. The first stage 
obtains the respective output scores of the different single classifiers described in Section~\ref{sec:Dago} based on different feature and patterns. 
The classifier in the second stage considers those scores as inputs.

We have performed an extensive experimental evaluation of possible combinations of those classifiers presented in 
Table~\ref{tab:Dagoresultsbasics}. Starting with a fusion of C1 and C2, the best final 
configuration fuses the information obtained from the following descriptors:

\begin{itemize}
 \item \textbf{C1}. HOG of the facial pattern (F-HOG).
 \item \textbf{C2}. HOG of the head and shoulders pattern (HS$_{64\times 64}$-HOG).
 \item \textbf{C3}. Concatenated LBP histogram extracted from the facial pattern (F-LBP$^{u2}$).
 \item \textbf{C4}. Concatenated LOSIB histogram extracted from the facial pattern (F-LOSIB).
 \item \textbf{C5}. Concatenated LOSIB histogram from the head and shoulders pattern (HS$_{64\times 64}$-LOSIB).
\end{itemize}

Each first stage classifier is trained using a SVM+RBF. The second stage feeds their respective scores into a new SVM+RBF classifier in charge of 
taking the final decision. The results achieved for the Dago's protocol are reported in Table~\ref{tab:Dagoensemble}. In order to confirm the influence 
of the combination in the results, a Kruskal-Wallis test was carried out and 
the difference in accuracy was found significant ($p=4.7\cdot 10 ^{-5}$). 
The ANOVA test was discarded because a previous Jarque-Bera test 
rejected the normality of the samples ($p=0.5$).

\begin{figure}[!t]  \begin{minipage}[t]{1.0\linewidth}
    \centering
   
    \begin{minipage}[c]{0.85\linewidth}
      \centering
      \includegraphics[width=1\linewidth]{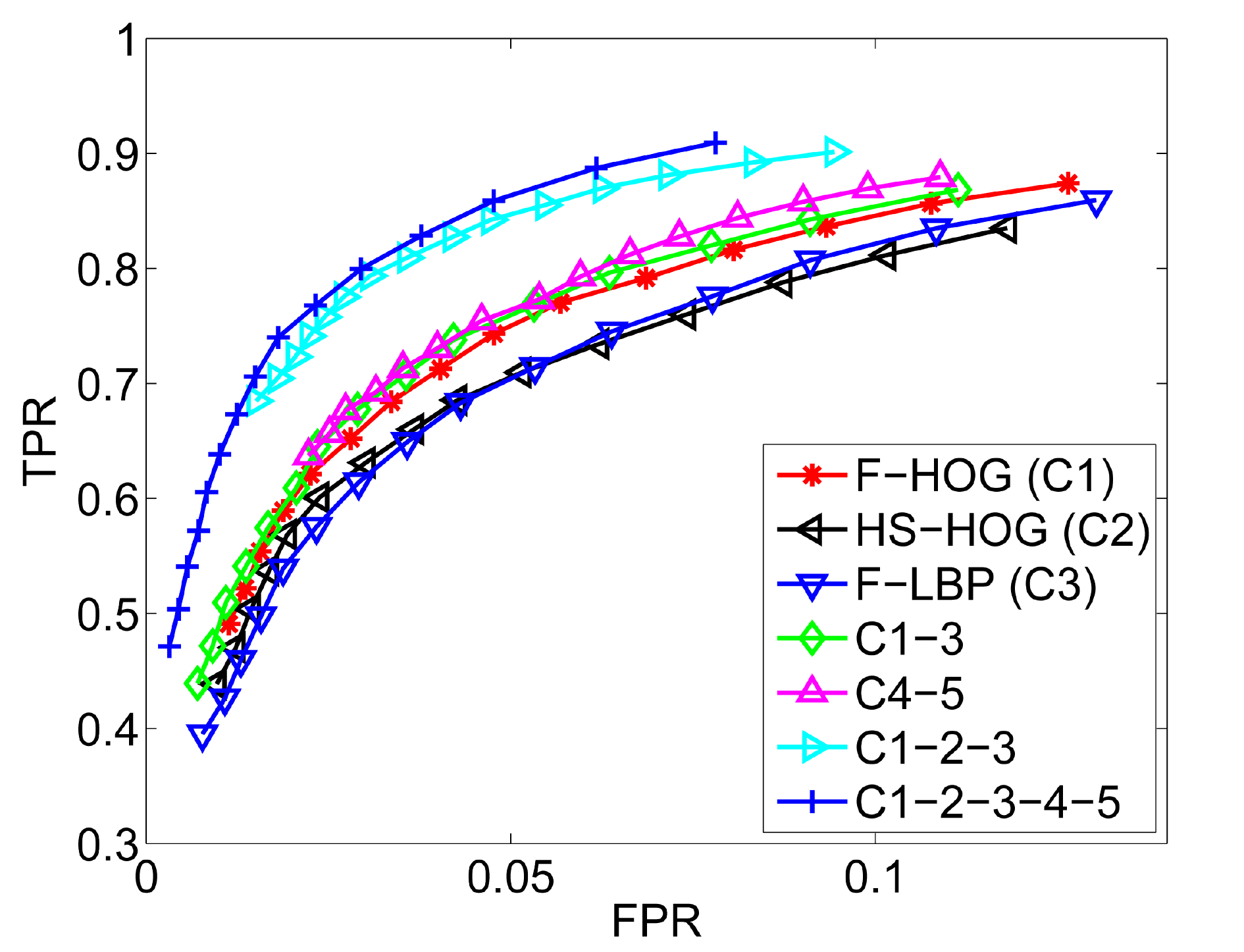}\\
       \end{minipage}\hfill 
 
\caption{ROC curves using the Dago's protocol. \label{fig:ROCs}}
  \end{minipage}
\end{figure}

The results confirm the 
initial hypothesis of the complementary information contained in the different descriptors. This is evident observing the fusion of features from F, e.g. S1 and S2. There is also a benefit when features are extracted from both F and HS, as evidenced in S3 where LOSIB features are extracted from both patterns, or in S4 and S5. The former fuses LBP and HOG features, the latter integrates also LOSIB features. 
The accuracy reaches $91.64\%$ (up to $94.28\%$ for adults), beating both the results 
previously presented in this work, and the literature
for this experimental setup. The 
observation of the respective ROC curves, see Figure~\ref{fig:ROCs}, confirms the best performance of S5.

\begin{table}[ht]
\caption{GC accuracy (in brackets female/male) using stacked classifiers for GROUPS.}%
\centering
\begin{footnotesize}
 \begin{tabular}{c|c|c|c}
  \multicolumn{2}{c|}{Classifiers fused} & \multicolumn{2}{|c}{Accuracy (\%)}\\ \hline  
   \multicolumn{2}{c|}{\multirow{2}{*}{C1-C3 (S1)}} & \multicolumn{2}{|c}{$88.59$}  \\  
  \multicolumn{2}{c|}{}& \multicolumn{2}{|c}{($88.09$/$89.10$)} \\
  \hline
  \multicolumn{2}{c|}{\multirow{2}{*}{C1-C3-C4 (S2)}} & \multicolumn{2}{|c}{$89.18$}  \\  
  \multicolumn{2}{c|}{}& \multicolumn{2}{|c}{($88.98$ /$89.38$)} \\
  \hline
  \multicolumn{2}{c|}{\multirow{2}{*}{C4-C5 (S3)}} & \multicolumn{2}{|c}{$88.72$}  \\  
  \multicolumn{2}{c|}{}& \multicolumn{2}{|c}{($87.67$/$89.79$)} \\ 
  \hline
   \multicolumn{2}{c|}{\multirow{2}{*}{C1-C2-C3 (S4)}} & \multicolumn{2}{|c}{$90.44$}  \\  
  \multicolumn{2}{c|}{}& \multicolumn{2}{|c}{($90.22$/$90.66$)} \\
  \hline
  
   \multicolumn{2}{c|}{\multirow{2}{*}{C1-C2-C3-C4-C5 (S5)}} & \multicolumn{2}{|c}{$91.65$}  \\  
  \multicolumn{2}{c|}{}& \multicolumn{2}{|c}{($91.05$/$92.26$)} \\
  \hline
   \multicolumn{2}{c|}{\multirow{2}{*}{C1-C2-C3-C4-C5 (adults)}} & \multicolumn{2}{|c}{$\textbf{94.28}$}  \\  
  \multicolumn{2}{c|}{}& \multicolumn{2}{|c}{($94.40$/$94.16$)} \\
\end{tabular}
\end{footnotesize}
\label{tab:Dagoensemble}
\end{table}

\begin{figure}[hbt]  
\begin{minipage}[t]{1.0\linewidth}
    \centering
   \centering
        \begin{minipage}[c]{0.33\linewidth}
      \centering
      \includegraphics[scale=0.17]{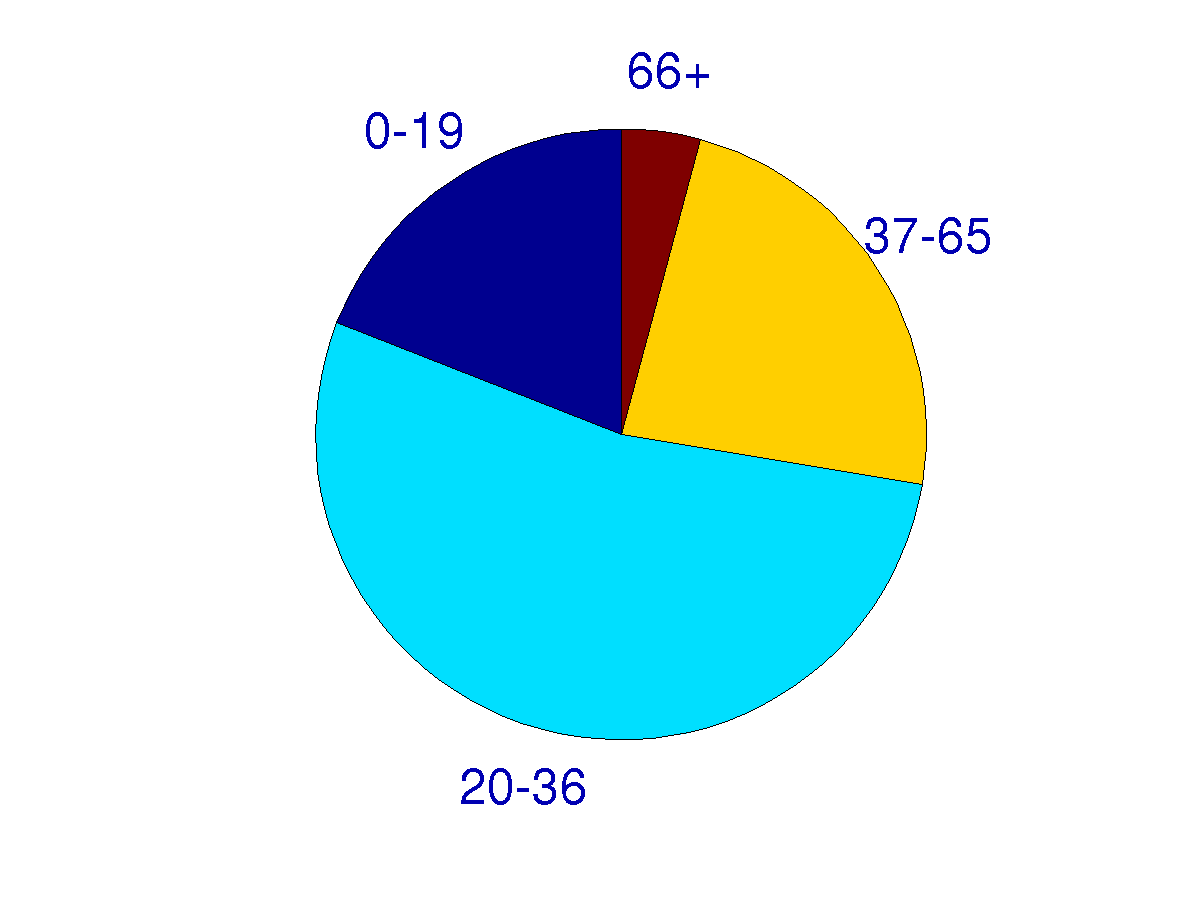}
    \end{minipage}\hfill 
     \begin{minipage}[c]{0.34\linewidth}
      \centering
      \includegraphics[scale=0.17]{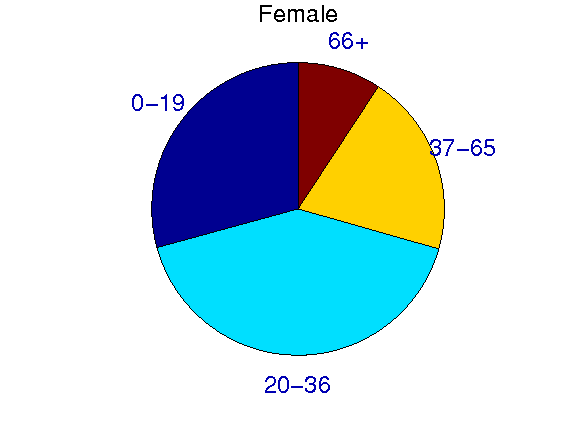}
    \end{minipage}\hfill
     \begin{minipage}[c]{0.33\linewidth}
      \centering
      \includegraphics[scale=0.17]{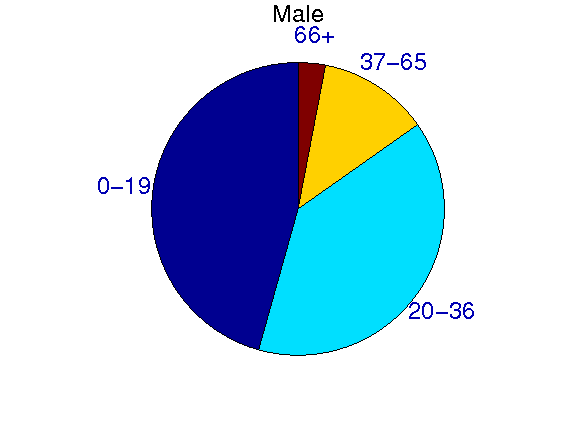}
    \end{minipage}\hfill \\
     \caption{Ratio (left) of samples per age range in GROUPS, distribution 
     (miggle and right) of errors per age range and class. 
     \label{fig:AgeErrors}}
  \end{minipage}
\end{figure}

The reader may have observed the large improvement when only adults ($>20$ years old) 
are considered in both training and test sets. As recently
analyzed, gender discriminant features in children differ from 
adults~\cite{Satta14-icpr}. This effect is illustrated in Figure~\ref{fig:AgeErrors}, 
where
the GROUPS samples age groups distribution is presented on the left, and the error 
per gender and age group distribution on the right. Both children and 
elderly affect negatively 
the overall recognition
accuracy. The former particularly among males, the latter particularly among females. 
In GROUPS the presence of children is much larger, therefore there is a 
larger improvement when they are not considered 
for both test and training, reaching over $94\%$ accuracy.

\begin{figure}[!t]  \begin{minipage}[t]{1.0\linewidth}
    \centering
   
    \begin{minipage}[c]{1\linewidth}
      \centering
      \includegraphics[width=0.85\linewidth]{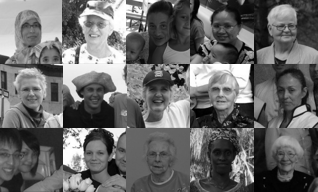}\\
       \end{minipage}\hfill 
 
\caption{Examples of female samples (HS) with no correct classification. \label{fig:Errores}}
  \end{minipage}
\end{figure}

The influence of elderly is reflected in Figure~\ref{fig:Errores}. We present there samples that were incorrectly classified
by all the first stage classifiers. Observing the female failures in detail,
there is a large presence of elderly ladies, suggesting the inadequate modeling of that particular appearance. Other errors seem to be related
with the presence of both glasses, and hats or other elements. The former is affecting the facial features, the latter blocks what may be coped with the local 
context.

\begin{figure}[!t]  \begin{minipage}[t]{1.0\linewidth}
    \centering
   
    \begin{minipage}[c]{0.85\linewidth}
      \centering
      \includegraphics[width=1\linewidth]{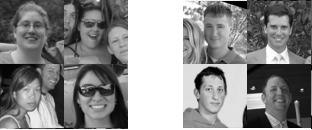}\\
       \end{minipage}\hfill 
 
\caption{Examples of female (left) and male (right) samples correctly classified after adding HS features. \label{fig:Aciertos}}
  \end{minipage}
\end{figure}

Finally, Figure~\ref{fig:Aciertos} presents samples (age range $20-36$) that were incorrectly classified using the combination S2, i.e. using only 
the facial features, but correctly classified with the full ensemble of classifiers, i.e. S5, using both facial and local context details. This set of 
ambiguous facial patterns was better described adding HS features.

\subsection{In- and cross-database results in full datasets}

In a final experiment, we have tested the best performing classifier in the full selected databases: LFW, GROUPS and MORPH. The in-database (highlighted) 
and cross-database 
results are presented in the upper half of Table~\ref{tab:FullDBCross}. 

Starting with GROUPS, a slight decrease, compared to the 
Dago's protocol, is observed. Certainly, the inclusion of the whole dataset introduces low resolution and non automatically detected faces, circumstance
that adds challenging aspects in the experiment. However, the accuracy is closer to $91\%$, and to $94\%$ if only adults are
considered for training and test. Those rates beat any previous results
in the dataset, but the best is still to come. 

For the other two datasets: MORPH and LFW. We already mentioned that the state-of-the-art literature reports $97-98\%$ accuracies. We achieved that accuracy 
for MORPH, but \textit{only} $95\%$ for LFW. We argue that as stated in those 
works~\cite{Tapia13-ifs,Shan12-prl}, the authors skipped \textit{faces that are not (near) frontal}, and used higher resolution. The comparison is therefore not completely as fair as it should.

\begin{figure}[htb]  
   \centering
   \includegraphics[height=60mm]{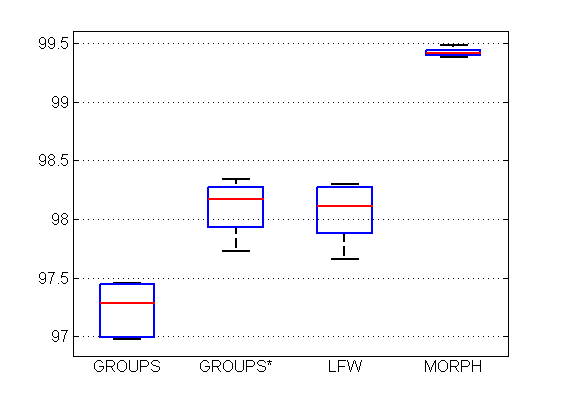}
   \caption{Boxplot of the accuracy for in-database results (GROUPS* $>20$ years old).}
   \label{fig:boxplot-accuracy}
\end{figure}

Focusing on the most challenging cross-database performance, the observation of Table~\ref{tab:FullDBCross} evidences firstly an improvement compared to 
previous literature, even if full datasets are evaluated.
This suggests that if a more challenging or general problem is carried out, 
the complementary information contained in both the descriptors, and the 
regions of interest helps, 
particularly to tackle large 
variability, and smaller image resolutions. This achievement is new if compared 
to the exclusively 
face centered classifiers.

However, we were not completely satisfied, and analyzed alternatives. 
CNN~\cite{Lecun98-ieee} have lately achieved relevant results in many 
Computer Vision problems as image classification~\cite{Krizhevsky12-nips}. 
In this sense, some authors have started to evaluate them in GC, with some results
reported for LFW and GROUPS~\cite{Mansanet16-prl,Antipov15-prl}. We have adopted 
the CNN design proposed by \cite{Levi15-amfg} with three 
convolutional layers and two fully connected layers, trained with HS pattern ($159 \times 155$ pixels), see section~\ref{sec:faces}.
The achieved results are summarized in the lower half of Table~\ref{tab:FullDBCross},
presenting in most cases slightly better accuracies particularly when 
testing with GROUPS.

\begin{table}[ht]
\caption{GC accuracy ($\%$) with full datasets using hand crafted (left) and CNN (right) classification. 
The table includes
in- (5-folds cross validation) and cross-database results. In-database results are highlighted. $^1$ $>20$ years old}%
\centering
\begin{small}
\begin{tabular}{c|c|c|c}
  \multirow{2}{*}{Training set}& \multicolumn{3}{c||}{Test set (hand crafted)} \\
  \cline{2-4}
  &  GROUPS & LFW & MORPH 
   \\  \hline
   GROUPS  &  \cellcolor{Gray} $90.85$& $94.10$ & $89.98$ \\
\hline
   GROUPS$^1$  &  \cellcolor{Gray} $93.89$& $93.94$ & $88.11$ \\  
\hline
   LFW &  $80.22$ & \cellcolor{Gray}$95$ & $89.16$ \\  
\hline
   MORPH   &  $62.04$&  $84.53$ & \cellcolor{Gray}$98.85$ \\ 
\hline \hline
\multirow{2}{*}{Training set} & \multicolumn{3}{c}{Test set (CNN)}\\
  \cline{2-4}
  &  GROUPS & LFW & MORPH 
   \\  \hline
   GROUPS  &\cellcolor{Gray} $92.90$& $94.48$ & $87.56$\\
\hline
   GROUPS$^1$  &  \cellcolor{Gray} $95.82$& $94.64$ & $90.80$\\  
\hline
   LFW & $85.92$ & \cellcolor{Gray}$96.7$ & $91.84$\\  
\hline
   MORPH    &  $72.32$&  $83.16$ & \cellcolor{Gray}$98.77$\\ 

\end{tabular}
\end{small}
\label{tab:FullDBCross}
\end{table}

\begin{table}[ht]
\caption{GC accuracy ($\%$) combining hand crafted features and CNN.}%
\centering
\begin{small}
\begin{tabular}{c|c|c|c}
  \multirow{2}{*}{Training set}& \multicolumn{3}{c}{Test set} \\
  \cline{2-4}
  &  GROUPS & LFW & MORPH 
   \\  \hline
   GROUPS  &  \cellcolor{Gray} $97.23$& $98.00$ & $93.46$ \\
\hline
  GROUPS$^1$  &  \cellcolor{Gray} $98.10$& $97.95$ & $92.98$ \\  
\hline
   LFW &  $90.14$ & \cellcolor{Gray}$98.06$ & $93.54$\\  
\hline
   MORPH   &  $67.40$&  $88.70$ & \cellcolor{Gray}$99.42$\\ 
\end{tabular}
\end{small}
\label{tab:FullDBCNNCross}
\end{table}

This observation, added to some promising very recent results combining CNN and 
hand crafted 
features~\cite{Mansanet16-prl,Wolfshaar15}, have guided us to combine
local descriptors and CNN, integrating in the proposal another first stage classifier, 
this time based on CNN. The achieved results are summarized in 
Table~\ref{tab:FullDBCNNCross}. The new results evidence that with the exception of 
one situation,
the one with originally lowest accuracy (MORPH for training, GROUPS for testing), the proposed combination 
 boosted all the accuracies remarkably.

Reviewing first the in-dataset 5-fold cross-validation evaluations, LFW and MORPH reported respectively 
98.06\% and 99.42\%. Thy are indeed new state-of-the-art for both, but
certainly similar rates have been achieved for the LFW subset containing 
almost frontal faces~\cite{Tapia13-ifs,Shan12-prl}.
The accuracy achieved for GROUPS boosted up to 97.23\% and to 98.1\% when only faces
over 19 were used for both training and test. We have no previous referece of any 
similar reported accuracy neither for the whole dataset, or a subset.

Considering cross-database, training with GROUPS and testing with LFW reported
similar
numbers to LFW in-dataset results. In fact, previous state-of-the-art 
for cross-dataset with
LFW has already reported 97\%, but that was achieved 
training with 400,000 samples~\cite{Antipov15-prl} or four millions~\cite{JiaS15-prl}
Here that accuracy is beaten, reaching 98\%, and that is done with just a 7\% 
of the training samples used by Antipov et al. When training with
MORPH GC rates are lower, 88\%, but significantly better that recent reported 
results that reached $76\%$~\cite{Erdogmus14-mmsp}. On the other side, GROUPS present larger
difficulties, being extremely complex if training with MORPH, just $67\%$, and easier
training with LFW. Again the achieved performance over $90\%$, is more than $5\%$ better
than the latest reported result by Danisman et al.\cite{Danisman15-esa}, i.e. $85\%$

In order to test the significance in performance in the datasets, a previous 
Jarque-Bera test was carried out to assess the normality of the samples and 
the results was that the samples are not normally distributed ($p=0.5$) so a 
Kruskal-Wallis test was conducted instead of the ANOVA test. As result, it is 
found a significant difference ($p=1.1\cdot 10 ^{-3}$) in the performance of 
the datasets, see Figure \ref{fig:boxplot-accuracy}.




\subsection{Discussion}

In short, we have compared two different focuses to tackle the GC problem.
On the one side, we have made use of previous computer vision experience to setup
a solution based on local descriptors, that required the almost manual exploration of alternatives, configuration setups, and areas of interest. This 
focus is now being referred as hand crafted features based, due to the high cost
given to the system designer to select the parameters. On the other side, to translate the feature selection and tuning work load to the training process, deep-CNN have also been studied within the same
experimental setup.

The evaluation of both solutions reported similar performances, 
even if a closer look
indicates a tiny advantage in terms of accuracy for CNN, compare left and fight 
accuracies in Table~\ref{tab:FullDBCross}.

However, we have proven that the combination at score level of both approaches 
reaches higher levels of GC accuracy, evidencing better performance 
both for in- and cross-database evaluations, suggesting the different
complementary information for GC provided by descriptors, regions of interest and CNN.
For cross-database, 
previous reported accuracies were typically lower than in-database accuracies. 
We have however confirmed the claim made by Klare et al.~\cite{Klare10-btas}
suggesting the importance of demographic variety in the training data. In fact training with GROUPS and testing with LFW reached similar marks than the in-database evaluation for LFW. Under our knowledge, this is the first time that this fact has been made with a relative reduced number of samples.

Observing the accuracies achieved for in the wild scenarios, we may
wonder whether the problem is solved. Certainly, there is no real evidence
that rates over $97-98\%$ will be kept after deployment in real world conditions. 
However,
we consider that it is the right time to think about building more challenging
in the wild benchmarks to evidence existing difficulties.
A very recent 
work focused on the MORPH 
dataset~\cite{Guo14-ivc} presents a joint estimation framework, 
dealing with the influence of gender and ethnicity on age estimation. 
Their in-database results achieved 
high rates for GC ($95-98\%$) in MORPH. However, we are concerned 
with this particular kind of databases that certainly presents
a selection bias containing several samples of the same individual. 
For this reason, we have studied the GC errors in GROUPS. Those age ranges less 
present produce more classification errors using local descriptors, while their influence is irrelevant for CNN. However, removing under 20 years
old, affects positively both approaches as the population appearance seems to be less spread.
 
\begin{figure}[ht]
\caption{Mean facial patterns per gender and age group in GROUPS.}\label{GROUPSmeans}
\centering
\begin{tabular}{m{1cm}CCCC}
& 0-19 & 20-36 & 37-65 & 66+ \\
Female &  
\includegraphics[scale=0.8]{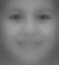}   &
\includegraphics[scale=0.8]{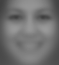}   &
\includegraphics[scale=0.8]{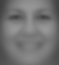}   &
\includegraphics[scale=0.8]{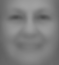}   \\
Male &
\includegraphics[scale=0.8]{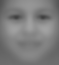}   &
\includegraphics[scale=0.8]{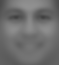}   &
\includegraphics[scale=0.8]{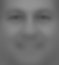}   &
\includegraphics[scale=0.8]{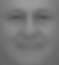}   \\
\end{tabular}
\end{figure}

In any case, GROUPS is in fact not ideal, there is a bias in the dataset, that may 
be illustrated observing the mean face images per age group and gender, see
Figure~\ref{GROUPSmeans}.  The average faces are smiling, suggesting the 
particular capture conditions used for the image collection.

Assuming newer and more challenging datasets, we may think about future lines 
of improvement, 
that  for our proposal may focus on the CNN architecture design, the features and areas of interest 
analyzed, or even the fusion approach used. In any case, two different 
GC architecures may be considered as suggested by Klare et al.~\cite{Klare10-btas}:
 a single gender classifier able to handle each demographic group, or a group of
 classifiers that may be tuned for each demographic group.


\section{Conclusions}

In this work, we have extensively explored the use of several descriptors and areas of interest for the face based GC problem. The ensemble of different 
classifiers considering a set of local descriptors and regions of interest in a stacking fashion has proven to reduce almost $20\%$ the previous error 
rate in the challenging GROUPS dataset following the Dago's protocol. 

These results were later confirmed in an experimental evaluation considering in- and cross-database classification in three large datasets: GROUPS, LFW and 
MORPH. Firstly, the in-database experiment with GROUPS keeps quite similar classification rates, over to $94\%$ in adults. The other two datasets reported 
similar accuracy rates to the best recent literature.

The experimental evaluation was carried out also for a CNN. The comparison 
indicates that both approaches perform similarly, with slight advantage 
for CNN in some experimental scenarios. A further exploration fused 
both focuses
in a score level combination.
The accuracies in full in-database cross-validation evaluations for 
GROUPS, LFW and MOPRH
were respectively  boosted up to over $97\%$ ($98\%$ in adults), $98\%$ 
and $99\%$, reporting also new state-of-the-art accuracies 
in cross-database GC performance.

\section*{Acknowledgments}
Work partially funded by the project TIN2015 64395-R from the Spanish Ministry
of Economy and Competivity, the Institute of Intelligent Systems and Numerical 
Applications in Engineering (SIANI) and the Computer Science Department at ULPGC.

\bibliographystyle{IEEEbib}
\bibliography{bibliografia_gias}

\end{document}